\pdfoutput=1

\documentclass[11pt]{article}

\usepackage[most]{tcolorbox}
\usepackage{authblk}
\usepackage{xcolor}
\usepackage{lmodern}
\usepackage{hyperref}
\usepackage{tabularx}

\usepackage{utfsym}

\usepackage[final]{ACL2023}
\usepackage{multirow}
\usepackage{times}
\usepackage{latexsym}
\usepackage{booktabs}
\usepackage{algorithm} 
\usepackage{algpseudocode} 
\usepackage{natbib}

\usepackage{pgfplots}
\pgfplotsset{compat=1.18}

\usepackage{amsmath}
\usepackage[T1]{fontenc}

\usepackage[utf8]{inputenc}

\usepackage{graphicx} 

\usepackage{microtype}

\usepackage{inconsolata}
\usepackage{float}

\definecolor{titlebar}{RGB}{50, 50, 50} 
\definecolor{textcolor}{RGB}{0, 0, 0}

\title{\includegraphics[width=1cm]{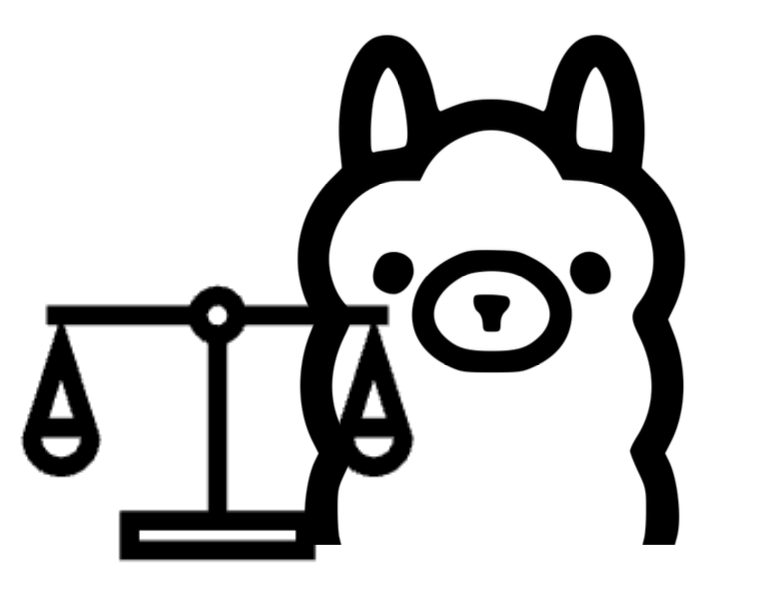} ASP2LJ : An Adversarial Self-Play Laywer Augmented Legal Judgment Framework }

\author{
  \textbf{Ao Chang}$^{1,2}$ \quad \textbf{Tong Zhou}$^{1,2}$ \quad \textbf{Yubo Chen}$^{1,2}$ $\thanks{*Corresponding author}$  \\
  \textbf{Delai Qiu}$^{3}$ \quad \textbf{Shengping Liu}$^{3}$ \quad \textbf{Kang Liu}$^{1,2}$ \quad \textbf{Jun Zhao}$^{1,2}$ \\

  $^1$The Key Laboratory of Cognition and Decision Intelligence for Complex Systems, \\
  Institute of Automation, Chinese Academy of Sciences, Beijing, China\\
  $^2$School of Artificial Intelligence, University of Chinese Academy of Sciences, Beijing, China\\
  $^3$Unisound, Beijing, China\\
}

\begin{document}
\maketitle

\begin{abstract}
Legal Judgment Prediction (LJP) aims to predict judicial outcomes, including relevant legal charge, terms, and fines, which is a crucial process in Large Language Model(LLM). However, LJP faces two key challenges: (1)Long Tail Distribution: Current datasets, derived from authentic cases, suffer from high human annotation costs and imbalanced distributions, leading to model performance degradation. (2)Lawyer's Improvement: Existing systems focus on enhancing judges' decision-making but neglect the critical role of lawyers in refining arguments, which limits overall judicial accuracy. To address these issues, we propose an \textbf{A}dversarial \textbf{S}elf-\textbf{P}lay \textbf{L}awyer Augmented \textbf{L}egal \textbf{J}udgment Framework, called ASP2LJ, which integrates a case generation module to tackle long-tailed data distributions and an adversarial self-play mechanism to enhance lawyers' argumentation skills. Our framework enables a judge to reference evolved lawyers' arguments, improving the objectivity, fairness, and rationality of judicial decisions. Besides, We also introduce RareCases, a dataset for rare legal cases in China, which contains 120 tail-end cases. We demonstrate the effectiveness of our approach on the SimuCourt dataset and our RareCases dataset. Experimental results show our framework brings improvements, indicating its utilization. Our contributions include an integrated framework, a rare-case dataset, and publicly releasing datasets and code to support further research in automated judicial systems.
\end{abstract}

\begin{figure}[t]  
	\centering  
	\includegraphics[width=8cm, height=8cm]{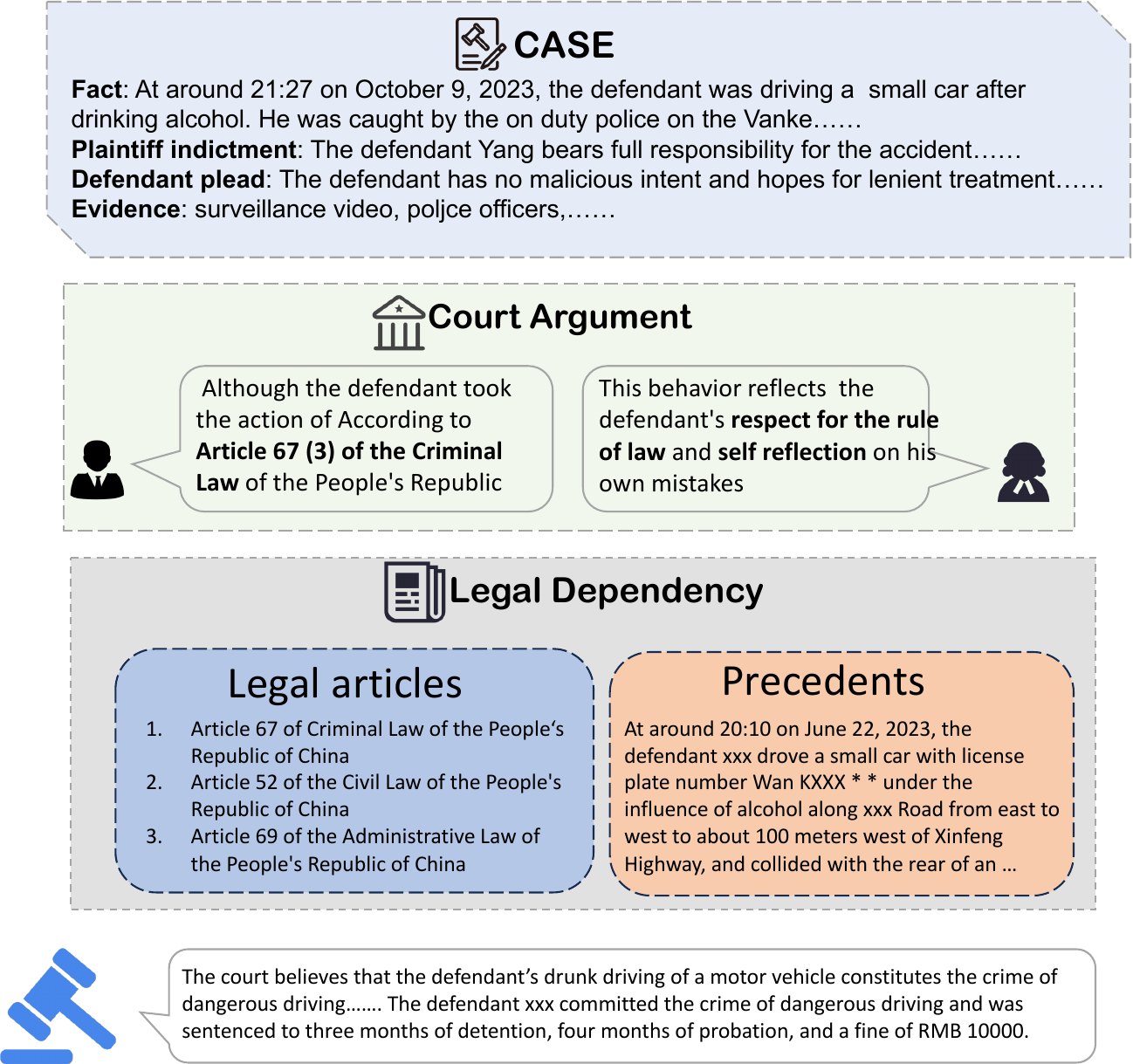}  
	\caption{Based on a real case, the lawyers argue and the judge make judgment according to legal dependency. }
	\label{fig1}
\end{figure}

\section{Introduction}
LJP task aims to predict relevant legal articles, prison terms, fines, and other judgment results of a legal case \citep{cui2022surveylegaljudgmentprediction}. With the advancement of deep learning, an increasing number of studies have been proposed to improve models' judgment prediction capacities \citep{xu2020distinguish, semo-etal-2022-classactionprediction}.
In recent years, the emergence of LLMs \citep{bai2023qwentechnicalreport, openai2024gpt4technicalreport} has made further progress in this field \citep{li-etal-2025-legal, wu2023precedentenhancedlegaljudgmentprediction}. LLMs can simulate a real legal courtroom and help improve the performance of LJP tasks \citep{feng2022legal, huang-etal-2024-cmdl}.
As illustrated in Figure \ref{fig1}, in real-world legal practice, the lawyers start to argue based on a provided case, and the judge should reference law articles and precedents to make a final judgment \citep{zhong2018legal, Ma_2021}.
 In the case law system \citep{mistica2020information}, judicial precedents can serve as binding authorities for court decisions, whereas in the civil law system \citep{buttner-habernal-2024-answering}, precedents are often utilized for interpretative guidance or reference purposes. 

However, despite these advancements, the current LJP task faces two major problems: cases' long tail distribution and the lawyer's improvement. 

\begin{table}[t]
    \centering
    \begin{small}
    \begin{tabular}{l|c|c|c|c}
        \toprule
        data & sum & criminal & civil  & admin \\
        \midrule
        cases & 27.3M & 1.8M & 24.7M & 0.68M \\
        laws & 13,117 & 5,425 & \multicolumn{2}{c}{7,692} \\
        \bottomrule
    \end{tabular}
    \end{small}
    \caption{Case Volume of our corpus}
    \label{table9}
\end{table}

\textbf{Case Distribution}: The distribution of the data adheres to the 80/20 rule, and some rare cases are paid less attention by LLM \citep{Li2025AddressingLD}, which results in the degradation of the model. As illustrated in Figure \ref{Figure2}, the current distribution of cases exhibits a long-tail characteristic \citep{refId0}, meaning that certain types of cases represent a very small proportion of the overall cases, and a human judge is puzzled when he meets such cases. Empirical studies \citep{garrett2011convicting, gross2014rate, zavrvsnik2021algorithmic} demonstrate that specific cases have a higher probability of being overturned, such as contract disputes, sexual assault, murder, etc.
For automated judicial systems, the issue of data distribution leads to a lack of generalizability in models when dealing with rare cases. As indicated in Table \ref{table4}, even state-of-the-art models exhibit performance degradation when processing infrequent cases compared to more common ones. While \citet{wang2024crosslingualstatutoryarticleretrieval} attempts to address this issue through LLM-generated cases, these methods still require manual judgment annotation, limiting their scalability and practical applicability. These limitations highlight the need for improved approaches to improve the capability of the automated judicial system to handle rare cases without human effort.

\textbf{Lawyer's Improvement}: Recently, some works \citep{he-etal-2024-agentscourt, chen2024agentcourtsimulatingcourtadversarial} try to introduce a simulated court to help the judge improve the judgment's accuracy. However, they focus on the judge while the lawyers don't get full improvement, which limits the performance of the final judgment. The role of legal professionals, particularly lawyers, has a crucial effect in achieving accurate judicial outcomes. Research findings \citep{poppe2015lawyers, anderson2012much, shiu1983role} reveal that seasoned legal practitioners achieve a greater success rate in litigation compared to their less experienced colleagues. Through their arguments, lawyers can present case analysis and relevant legal references, provide comprehensive perspectives, and identify potential points of contention, which helps the judge achieve a more transparent understanding of the case, leading to the pronouncement of a fair and just verdict.. According to the arguments, the judge can make an accurate judgment. Empirical studies \citep{habernal2024mining, sheppard2012sake} have demonstrated that judicial outcomes are influenced by the quality of legal arguments presented, leading to more equitable rulings. Nevertheless, current research in this domain faces limitations, either neglecting the role of legal arguments or being constrained by insufficient real-world data for optimizing legal argumentation. Therefore, enhancing the capabilities of lawyers to provide valuable references for judicial decision-making presents another significant challenge.

\begin{table}[t]
\centering
\scalebox{0.85}{
    \begin{tabular}{lcc}
    \toprule
    \textbf{Dataset} & \textbf{SimuCourt} & \textbf{RareCases} \\
    \midrule
    cases & 420 & 120 \\
    average articles & 3.71 & 6.90 \\
    if-rare? & \usym{2717} & \usym{2714} \\
    Average length per case & 440.9 & 384.7 \\
    \bottomrule
    \end{tabular}
}
\label{table1}
\caption{Basic statistics of the datasets.}
\end{table}

Motivated by these problems above, we propose an Adversarial Self-Play Laywer Augmented Legal Judgment framework, which enables the judge to reference the augmented lawyers' arguments and improve the objectivity, fairness, and rationality of judicial decisions of the judgment. 
In order to address the issue of real cases' long-tail distribution, we propose a case generation module. As the generated cases contain only case facts, plaintiffs' indictments, and defendants' pleas, excluding final judgments, our approach is independent of human judgment. It incorporates a case-court pipeline to mitigate the challenges posed by the long-tailed case distribution and facilitate the accumulation of judicial experience. 
To demonstrate the effectiveness of our method, we introduce a dataset called RareCases which encompasses rare cases in China. All the cases are sampled from the China Judgements Online. \footnote{https://wenshu.court.gov.cn}
Furthermore, in order to enhance the proficiency of lawyers, we propose an adversarial self-play mechanism for lawyer agents where the plaintiff and defendant lawyers engage in case analysis and confrontation, iteratively accumulating agent experience and improving their legal analysis capabilities. 
The system integrates lawyers' argumentative content with judicial decision-making modules to support more objective, impartial, and reasonable adjudication. 
The experimental results demonstrate that our framework effectively enhances the capabilities of the agents and exhibits strong performance across both datasets, particularly on RareCases.

Our main contributions are:
\begin{itemize}
\item[$\bullet$] We propose ASP2LJ framework, which is the first to incorporate a lawyer's perspective and utilize lawyers in a self-play optimized process for judgment.
\item[$\bullet$] We introduce RareCases, a legal dataset including the main rare cases, which provides an approach to assess the legal capacity of current LLMs.
\item[$\bullet$] We demonstrate the effectiveness of our framework by conducting experiments on SimuCourt, a public data set in China. Experimental results show that our framework outperforms the existing methods in various aspects. Impressively, in legal article generation, we get a 8\% higher than GPT-4 in recall score, indicating the utility of the proposed framework in LJP tasks.
\footnote{To enable further research, we will release our
datasets and code publicly.}
\end{itemize}

\begin{table}[t]    
    \centering
    \begin{small}
    \begin{tabular}{l|ccc}
        \toprule
        cases & Correctness & Reality & Rationality \\
        \midrule
        Cases sampled 500 & 0.99 & 0.99 & 0.98 \\
        \bottomrule
    \end{tabular}
    \end{small}
    \caption{Quality evaluation of our generated cases}
    \label{table8}
\end{table}

\begin{figure*}[t]  
	\centering  
	\includegraphics[width=12cm, height=8cm]{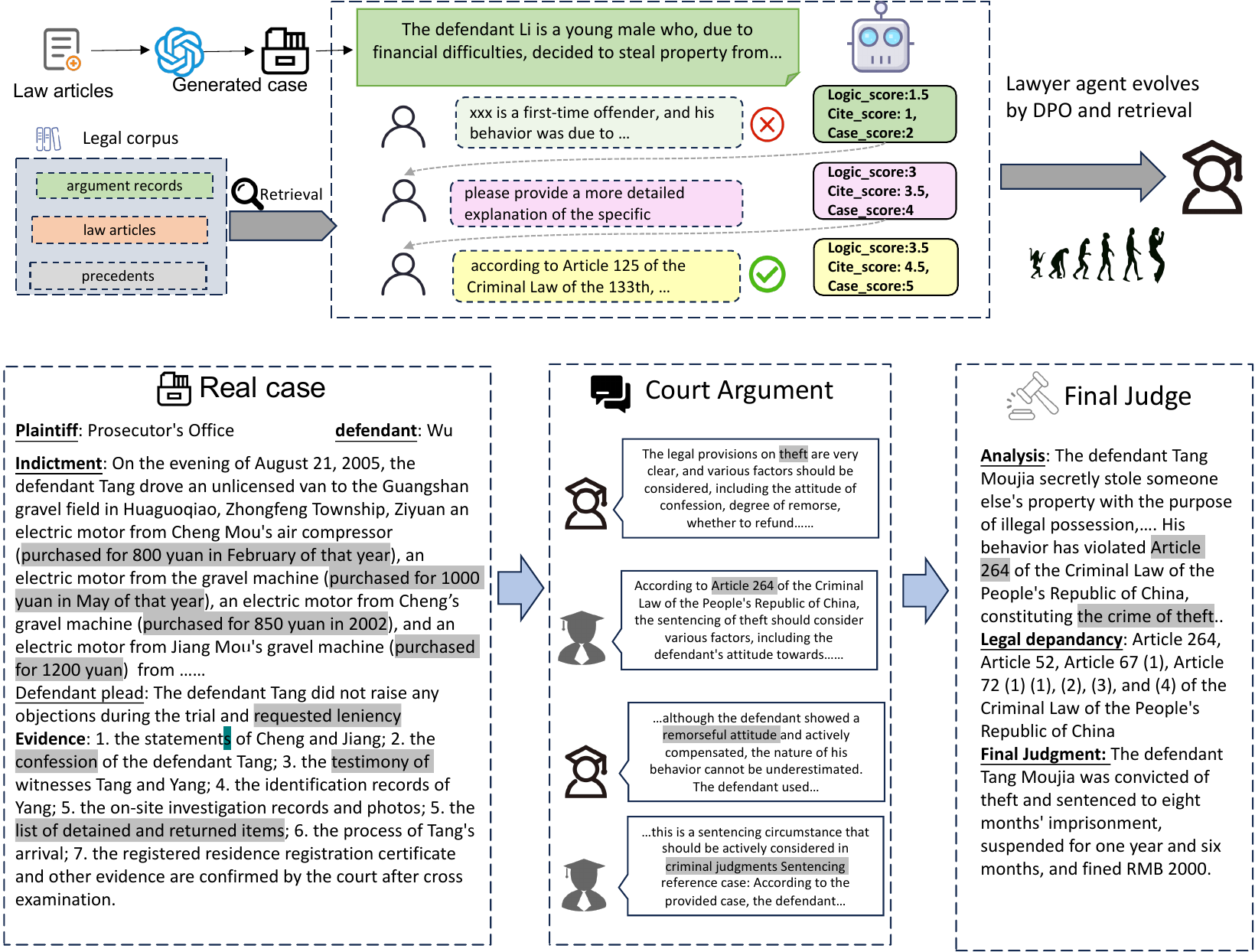}  
	\caption{Lawyer Evolution: Before a lawyer's speech, he will retrieve some corpus; after that, an evaluator scores and give some explanations to improve his speech}
	\label{fig2}  
\end{figure*}

\section{Method}
We propose an adversarial self-play lawyer augmented legal judgment framework that combines the adversarial argument process with the continuous enhancements of lawyer capabilities and the judge's Legal Judgment Prediction. This framework is designed to provide the judge with a more comprehensive understanding of cases, facilitating his impartial and accurate decision-making. The conceptual structure of this approach is illustrated in Figure \ref{fig2}.

\subsection{Court Pipeline}
To enhance the capabilities of lawyers within our framework, we implement a multi-stage approach that utilizes LLM to simulate court proceedings. Our pipeline is structured into three distinct phases: (1) legal case generation, (2) court argumentation, and (3) lawyer evolution, which aims to systematically improve the argumentation and reasoning skills of lawyer agents.

\subsubsection{Legal Case Generation}
Previous approaches rely on real cases as the foundation for legal arguments. However, these real cases' distribution follows a long-tail distribution, limiting LLMs' generalization. Lawyer agents also need diverse and adequate cases to enable their argumentation skills.
To overcome these limitations, we propose a pipeline that utilizes LLM to generate simulated legal cases automatically.

We use a vast collection of Chinese legal articles (criminal, civil, administrative). For each case generation, several legal articles are randomly selected to serve as the legal foundation, and GPT-4o is instructed to generate cases consisting of case facts, plaintiff's indictment, and defendant's pleadings. To ensure quality and complexity, we employ rejection sampling, discarding overly simple or anomalous cases after GPT-4o evaluation. This guarantees sufficiently complex and debatable cases, providing a strong platform for lawyers to hone their skills. Additionally, three law students verify the cases' correctness and logic." Detailed information is provided in Appendix \ref{sec:appendix B}.

This automated case generation process enables lawyer agents to engage in continuous argumentation grounded in a dynamic and expanding dataset. The increasing volume of generated cases provides lawyers with enhanced opportunities for expertise development, fostering progressive improvements in their professional capabilities. This approach not only mitigates the inherent limitations of relying exclusively on real-world case experience but also cultivates a dynamic environment conducive to the further evolution of lawyer agents.

\begin{table}[t]    
    \centering
    \begin{tiny}
    \begin{tabular}{l|ccccc}
        \toprule
        Model & Recall@100 & Recall@200 & Recall@500 & Recall@1000 \\
        \midrule
        BM25 & 0.055 & 0.106 & 0.221 & 0.296 \\
        LawFormer & 0.056 & 0.072 & 0.092 & 0.209 \\
        bge-m3 & 0.135 & 0.192 & 0.273 & 0.360 \\
        \textbf{ours} & \textbf{0.236} & \textbf{0.517} & \textbf{0.653} & \textbf{0.758} \\
        \bottomrule
    \end{tabular}
    \end{tiny}
    \caption{Degree of similarity between generated and real cases}
    \label{table2}
\end{table}

\subsubsection{Court Argumentation}
The argumentation process within the simulated court adheres to the procedural framework of real-world trials. Initially, the plaintiff is required to meticulously formulate the complaint, grounded in the case's factual description and relevant legal principles, clearly articulating the claims, factual basis, and legal reasoning. Conversely, the defendant's counsel must respond substantively to the complaint's content by addressing both factual determinations and legal applications, thereby constructing a comprehensive defense statement. Upon commencement of the formal argument phase, both parties engage in alternating presentations across three rounds, with each round comprising several core components: \\
\textbf{Statement} Both lawyers must articulate their own standpoints and legal claims. This serves as the foundation of the argument and the starting point for subsequent arguments. \\
\textbf{Retort} The plaintiff's lawyer must counter the arguments presented by the defendant in the previous round, pointing out logical flaws or errors in the legal application. The defendant's lawyer, in turn, must respond to the plaintiff's refutations while identifying weaknesses in the plaintiff's arguments. \\
\textbf{Legal Citations} When presenting their arguments, both lawyers must support their claims with relevant legal evidence, such as legal articles, judicial interpretations, and precedents. This not only tests their legal research skills but also their ability to extract relevant information and engage in logical reasoning. \\

Subsequent to each round of argumentation, GPT-4o offers lawyer agents an opportunity for reflective analysis and revision of their statements. The presiding judge evaluates lawyer performance across multiple dimensions, encompassing legal citation, logical reasoning, and factual description, and provides constructive feedback. Lawyers then leverage this feedback to refine their arguments, thereby enhancing the overall quality of their presentations. Through this multi-round, multi-faceted argument, the factual details of the case are fully revealed, and the points of contention are clearly presented. This process not only elevates the caliber of the lawyers' arguments but also facilitates the judge's ability to render an equitable verdict.

\begin{table*}[ht]
\centering
\begin{tiny} 
\begin{tabular}{lp{0.5cm}p{0.5cm}p{0.5cm}p{0.5cm}p{0.5cm}p{0.5cm}p{1cm}p{1cm}p{0.8cm}p{0.8cm}p{0.8cm}p{0.8cm}}
\toprule
\textbf{Model} & \multicolumn{3}{c}{\textbf{Legal Articles}} & \multicolumn{3}{c}{\textbf{Civil and Admini.}} & \multicolumn{3}{c}{\textbf{Criminal}} & \multicolumn{3}{c}{\textbf{Case Analysis}} \\
 & P & R & F & P & R & F & Charge & Term & Fine & Correct & Logic & Concise \\
\midrule
\multicolumn{12}{c}{First} \\
LawGPT & 0.113 & 0.060 & 0.072 & 0.215 & 0.436 & 0.300 & 0.690 & 0.139 & 0.092 & 0.332 & 0.468 & 0.390 \\
Qwen1.5-7B-Chat & 0.097 & 0.058 & 0.063 & 0.265 & 0.342 & 0.293 & 0.862 & 0.250 & 0.275 & 0.758 & 0.795 & 0.764 \\
GPT-3.5 & 0.130 & 0.115 & 0.118 & 0.335 & 0.418 & 0.332 & 0.850 & 0.487 & 0.312 & 0.677 & 0.714 & 0.710 \\
GPT-4 & \textbf{0.181} & \underline{0.133} & \underline{0.141} & \underline{0.458} & \underline{0.513} & \textbf{0.476} & \underline{0.881} & \textbf{0.531} & \underline{0.355} & \underline{0.835} & \underline{0.780} & \underline{0.835} \\
AgentsCourt & \underline{0.167} & 0.129 & 0.136 & \textbf{0.467} & 0.451 & \underline{0.453} & 0.875 & \underline{0.488} & 0.325 & 0.742 & 0.778 & 0.748 \\
ASP2LJ(ours) & 0.151 & \textbf{0.213} & \textbf{0.156} & 0.378 & \textbf{0.529} & 0.401 & \textbf{0.895} & 0.384 & \textbf{0.463} & \textbf{0.835} & \textbf{0.885} & \textbf{0.860} \\
\midrule
\multicolumn{12}{c}{Second} \\
LawGPT & 0.092 & 0.043 & 0.069 & 0.217 & 0.559 & 0.331 & 0.584 & 0.063 & 0.189 & 0.282 & 0.204 & 0.336 \\
Qwen1.5-7B-Chat & 0.204 & 0.058 & 0.085 & 0.29 & 0.65 & 0.38 & 0.860 & 0.098 & 0.228 & 0.390 & 0.425 & 0.425 \\
GPT-3.5 & 0.116 & 0.096 & 0.101 & 0.454 & 0.550 & 0.483 & 0.850 & 0.280 & 0.312 & 0.400 & 0.421 & 0.435 \\
GPT-4 & 0.171 & 0.161 & 0.161 & \textbf{0.729} & \textbf{0.754} & \textbf{0.737} & \underline{0.886} & \textbf{0.379} & 0.392 & \textbf{0.708} & \textbf{0.712} & \textbf{0.706} \\
AgentsCourt & \underline{}{0.226} & \underline{0.185} & \underline{0.189} & 0.549 & 0.617 & 0.571 & 0.849 & 0.320 & \textbf{0.509} & 0.463 & 0.526 & 0.476 \\
ASP2LJ(ours) & \textbf{0.271} & \textbf{0.223} & \textbf{0.231} & \underline{0.625} & \underline{0.697} & \underline{0.635} & \textbf{0.910} & \underline{0.363} & \underline{0.433} & \underline{0.586} & \underline{0.663} & \underline{0.640} \\
\bottomrule
\end{tabular}
\end{tiny} 
\caption{Overall performance of SimuCourt and baselines in the first and second instance experimental settings.}
\label{table3}
\end{table*}

\subsubsection{Lawyer Agent Evolution}
 The judge's understanding of a case is partially influenced by the argumentative skills of the legal representatives. How lawyers construct their arguments—through their form, structure, and content will significantly aid the judge in comprehending the case details. Effective legal representatives typically deliver arguments that are clear, logically structured, and supported by relevant legal citations and precedents, along with precise descriptions of the case's substantive relationships. Developing such skills necessitates ongoing practice and learning through simulated case scenarios.
 
 Our objective is to improve the argumentative proficiency of lawyers, enabling them to present case information in a more organized and comprehensive manner. To this end, we introduce a method for lawyer capacity enhancement that facilitates continuous learning and refinement of debating abilities, thereby enriching the data available for the judge's legal judgment prediction task. 

 Some works have demonstrated that judge's LJP capacity can improve by lawyers' arguments\cite{chen2024agentcourtsimulatingcourtadversarial, he-etal-2024-agentscourt, chen2025debatefeedbackmultiagentframeworkefficient}. However, they ignore that the lawyer agents can improve their argument abilities through agents' self-play evolution, which leaves space for further improvement. To address this gap, we introduce a subjective evaluation metric tailored to assess the quality of lawyers' arguments, to identify higher-quality presentations through a structured scoring system. The scoring framework focuses on three key dimensions: (1) the ability to accurately understand and cite relevant legal articles and precedents; (2) the logical coherence and organization of the argument; and (3) the depth and comprehensiveness of case analysis. Each dimension is scored on a scale of 0 to 5, yielding a total possible score of 15 points. In the first round, after each lawyer presents their argument, the content is evaluated using this metric, and constructive feedback is provided to guide improvements. The lawyer agent then refines its argument based on the feedback. This iterative process is repeated three times, and the highest-scoring argument is selected as the final submission. The detailed score criterion is illustrated in Table \ref{table10}.

\begin{table*}[t]
\centering
\begin{tiny}
\begin{tabular}{lcccccccp{0.8cm}cp{0.8cm}p{0.8cm}p{0.8cm}}
\toprule
\textbf{Model} & \multicolumn{3}{c}{\textbf{Legal Articles}} & \multicolumn{3}{c}{\textbf{Civil and Admini.}} & \multicolumn{3}{c}{\textbf{Criminal}} & \multicolumn{3}{c}{\textbf{Case Analysis}} \\
 & P & R & F & P & R & F & Charge & Prison term & Fine & Correct & Logic & Concise \\
\midrule
\multicolumn{12}{c}{} \\
LawGPT & 0.044 & 0.023 & 0.029 & 0.097 & 0.113 & 0.101 & 0.560 & 0.098 & 0.022 & 0.030 & 0.063 & 0.143 \\
Qwen1.5-7B-Chat & 0.053 & 0.041 & 0.042 & 0.082 & 0.101 & 0.090 & 0.759 & 0.107 & 0.030 & 0.133 & 0.197 & 0.247 \\
GPT-3.5 & 0.100 & 0.048 & 0.060 & 0.156 & 0.174 & 0.162 & 0.812 & 0.380 & 0.074 & 0.386 & 0.420 & 0.376 \\
GPT-4 & 0.147 & 0.115 & 0.120 & \textbf{0.182} & \textbf{0.198} & \textbf{0.186} & \underline{0.835} & \textbf{0.439} & \textbf{0.156} & \underline{0.426} & \underline{0.470} & \underline{0.447} \\
AgentsCourt & \underline{0.160} & \underline{0.118} & \underline{0.121} & 0.149 & 0.162 & 0.154 & 0.826 & \underline{0.417} & \underline{0.151} & 0.420 & \textbf{0.476} & 0.440 \\
ASP2LJ(ours) & \textbf{0.168} & \textbf{0.119} & \textbf{0.127} & \underline{0.160} & \underline{0.183} & \underline{0.167} & \textbf{0.850} & 0.161 & 0.107 & \textbf{0.433} & 0.447 & \textbf{0.520} \\
\midrule
\end{tabular}
\end{tiny}
\caption{Overall performance of our RareCases.}
\label{table4}
\end{table*}

\subsection{Judgment Prediction}
The lawyers who have evolved through simulated cases can engage in arguments on real cases, and the judge can reference the generated records. During the case judgment process, the judge not only considers the arguments presented by the lawyers but also utilizes an advanced legal retrieval system to search for relevant cases and legal articles. Therefore, we collect all the cases from Wenshu Web in 2021, encompassing criminal, civil, and administrative cases. The detailed information is in \ref{table9}. This retrieval mechanism ensures the precision and comprehensiveness of the LJP tasks. After completing the analysis of argument records and legal retrieval, the judge agent enters the legal judgment prediction phase. This phase mainly involves three core tasks: predicting the judgment results, determining the legal dependency, and analysing the whole case. 

\section{Experiment}
In this section, we start to evaluate the performance of our framework in downstream tasks. We will elaborate on our dataset, experiment design, and result analysis.
\subsection{Benchmark}
We adopt the SimuCourt benchmark of previous work, AgentsCourt\citep{he-etal-2024-agentscourt}, as the main part of our experiments. SimuCourt is a Chinese benchmark consisting of 420 cases that encompasses objective evaluations and subjective analyses, including first-instance and second-instance cases.

Besides, in order to evaluate the current models' capacity of handling rare cases, we propose our dataset called RareCases, which consists of 120 rare cases encompassing civil and criminal law. These legal cases are divided into "high", "mid", and "low" by their rarity, and "low" cases are rarer than "mid" and "high cases. Detailed information on its collection, processing, and verification is illustrated in the Appendix \ref{sec:appendix C}.

\subsection{Settings}

\textbf{Models}.
We adopt Qwen1.5-7B-Chat as base model and simultaneously compare it with lawGPT\citep{zhou2024lawgptchineselegalknowledgeenhanced}, GPT-3.5-turbo-0613, and GPT-4-1106-preview. Qwen1.5-7B-Chat is used as the base model for subsequent retrieval and optimization, and GPT-4o is instructed to evaluate and score the lawyer agents' arguments. \\

\textbf{Baselines}.
We compare our method with the following baselines: \\
(1) Vanilla. We choose Qwen1.5-7B-Chat, GPT3.5-turbo-0613, GPT-4-1106-preview as vanilla models. Besides, the base model of our framework is also Qwen1.5-7B-Chat.\\
(2) LawGPT\citep{zhou2024lawgptchineselegalknowledgeenhanced}. LawGPT is Chinese-LLaMA-7B fine-tuned on a dataset of 300,000 legal question-answer pairs.\\
(3) AgentsCourt\citep{he-etal-2024-agentscourt}. An LLM agent framework. They improve the judge's performance by introducing argument data and retrieving several law articles, precedents and law papers. We utilize GPT-3.5-turbo-1106 as its base model.\\

\textbf{Metric}.
We divide our task into three categories: legal articles, final judgment, and case analysis. For each task, we have specified some metrics to measure its performance. Detailed illustration is presented in Appendix \ref{sec:appendix D}.

\textbf{Finetune}. 
As shown in Figure \ref{fig2}, each statement is assessed and scored according to evaluation metrics. Every case undergoes three rounds of dialogue, with each round being evaluated three times. From these evaluations, we select the arguments with the highest and lowest scores to conduct DPO training. 
Besides, we instruct GPT-4o to generate 10,000 cases and fine-tune bge-m3.
Specifically, since our case generation is based on legal articles, each case is associated with several gold legal articles. For each case, we use BM25 to retrieve 50 law articles with the case fact as a query. The gold articles are deemed as positive samples, and other retrieved articles are negative samples. 
In this way, we can conduct DPO training and enhance the retriever's performance. 
As illustrated in Table \ref{table2}, our performance improves compared with other approaches. However, the peak performance achieved in Recall@1000 is only 75.8\%, indicating substantial scope for improvement in Legal Judgment Prediction (LJP) tasks.

\textbf{Retriever}. 
BGE-m3\citep{chen2024bgem3embeddingmultilingualmultifunctionality} is an advanced retriever proposed by BAAI, which leads to superior performance in multilingual retrieval, cross-lingual retrieval, and multilingual long document retrieval tasks, while in legal tasks, the sparse retriever BM25 \citep{rosa2021yesbm25strongbaseline} is in common use due to its relevance scoring algorithm. 
In this paper, we adopt a hybrid retrieval method to search for argument records, cases, and legal articles. Regarding argument records and cases, due to the context limitations of Qwen1.5-7B-Chat, we use BM25 for retrieval to obtain 100 candidate documents and then use BGE-M3 for reranking. Finally, 1 document is selected as the retrieved document. For legal articles, we use our fine-tuned bge-m3 to retrieve 200 legal articles as candidate articles.

\textbf{Corpus}. 
We collect all the Chinese legal cases in 2021, spanning criminal, civil and administrative cases, more than 27M as our case corpus. Each case in our corpus consists of five factors: case name, action cause, stage, relevant articles, and full text. To protect the privacy of involved parties, all case records were anonymized by removing personally identifiable information, including names, geographic locations, and other sensitive details. Detailed data statistics are shown in Table \ref{table9}.
Besides, we collect 13,117 law articles as our law article corpus. \\

\begin{figure}[t]
\centering  
\includegraphics[width=6cm, height=4cm]{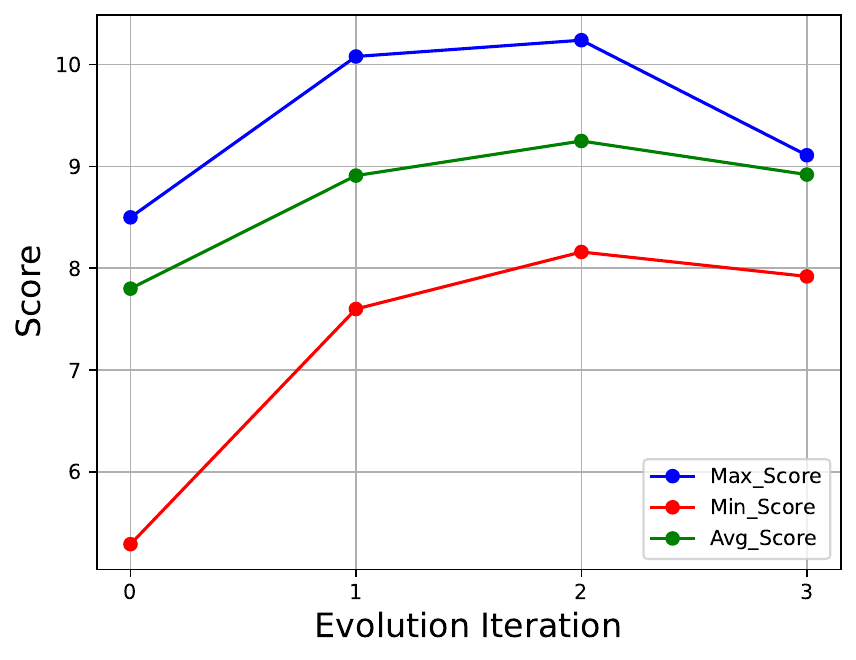}  
\caption{Score of lawyer agent by fine-tuning rounds}
\label{fig3}
\end{figure}

\subsection{Results}
Table \ref{table3} and Table \ref{table4} present the main experimental results on SimuCourt and our RareCourt, respectively.
\subsubsection{SimuCourt}
As Table \ref{table3} shows, our method outperforms other methods overall. Compared with the vanilla models, the performance has improved.\\ 
\textbf{Criminal Prediction} As for crime prediction, We extract the charge, prison term and fine by regular expressions. Among all the results, AgentsCourt achieves the best during the baselines, indicating the importance of argumentation and retrieval. Although our method brings large improvement, about 0.148 in criminal fine compared with vanilla Qwen-1.5-7B-Chat, we don't make obvious progress relative to GPT-4. There is still a large room for improvement.\\
\textbf{Civil and Administrative Prediction}. In the area of civil and administrative laws, our indicators comprehensively surpass those of the vanilla Qwen1.5-7B-Chat and GPT-3.5. However, GPT-4 still surpasses us by 10\% relatively. This also reflects that there is still a huge gap between current judgment and evaluation work in civil and administrative laws. \\
\textbf{Law Articles}. It is seen that the average accuracy of charge prediction by our method exceeds that of the vanilla Qwen1.5-7B-Chat by 15\%, and exceeds that of GPT-3.5 and GPT-4 by 10\% and 8\% respectively. However, the highest score is just 23.1\% in F1, which indicates that our agent framework needs further improvement.

\subsubsection{RareCases}
As shown in Table \ref{table4}, in prison term and fine tasks, ASP2LJ's score are much lower than GPT-4, GPT-3.5, and AgentsCourt, which shows that maybe Qwen1.5-7B-Chat is not sensitive to numbers. Compared to SimuCourt, the performance totally declines in RareCases. For example, in civil and administrative tasks, the highest score is just 18.6\%, which is much lower than 47.6\%, which leaves large room to advance. It is obvious that all the models perform worse in our RareCases, and the long-tail distribution truly obstructs the development of legal AI.

\begin{table}[t]
\centering
\begin{small}
\begin{tabular}{lccc}
\toprule
\textbf{Model} & \multicolumn{3}{c}{\textbf{Legal Articles}} \\
 & P & R & F  \\
\midrule
no argument & 
0.66 & 0.20 & 0.30 \\
Qwen1.5-7B-Chat & 0.76 & 0.23 & 0.34 \\
GPT-4o & \textbf{0.76} & \textbf{0.26} & \textbf{0.37}  \\
\midrule
\end{tabular}
\end{small} 
\caption{Randomly sample 60 cases' legal articles generation performance between different arguments generated by different models}
\label{table5}
\end{table}

\subsection{Ablation and Analysis}
\begin{figure}[t]
	\centering  
	\includegraphics[width=7cm, height=5cm]{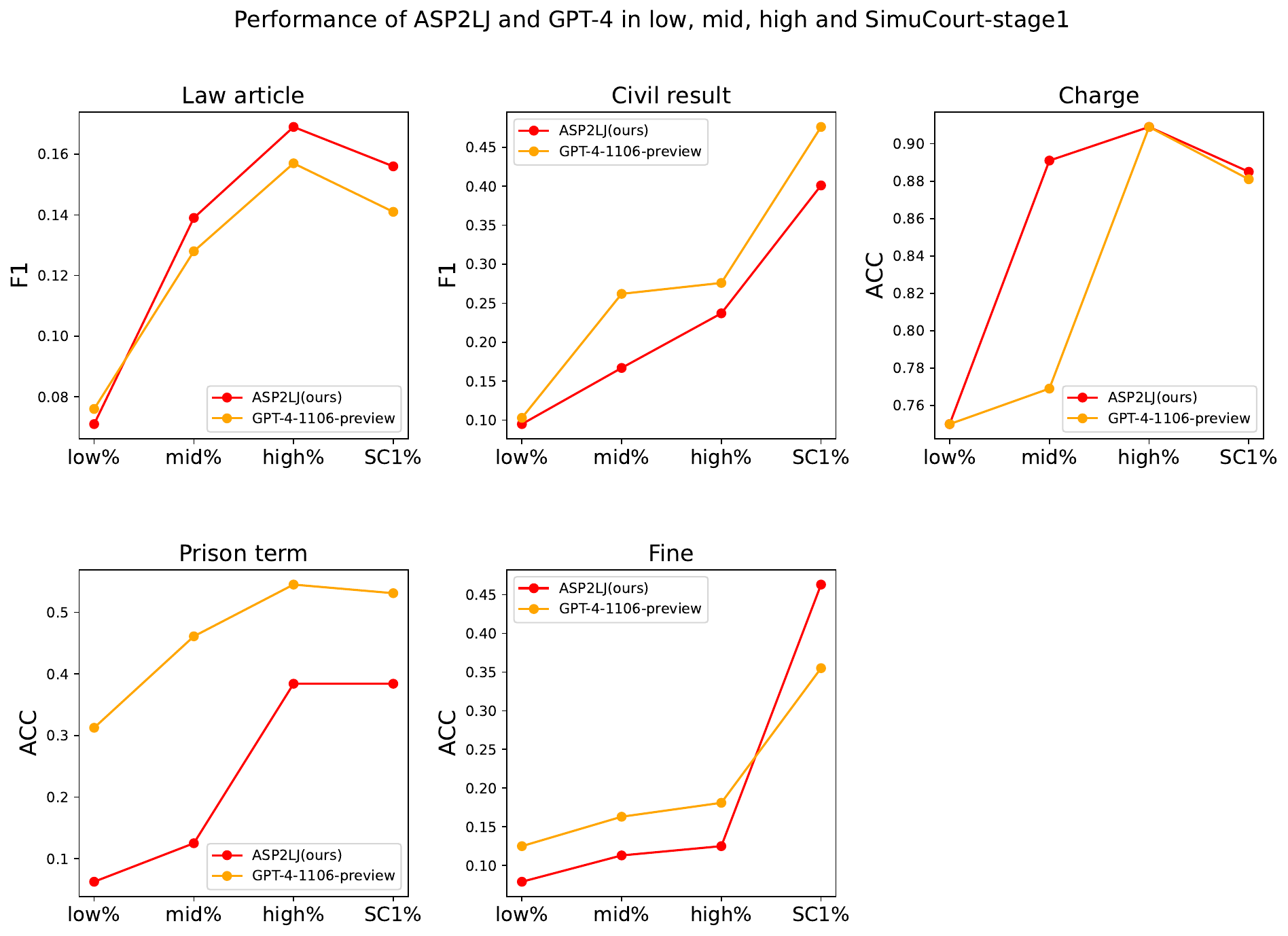}
	\caption{Models' performance between rare cases and common cases. Rare cases are divided into low, mid, and high, which represents their rarity. }
	\label{fig4}
\end{figure}

\textbf{Ablation} As demonstrated in the table \ref{table7}, retrieval plays a pivotal role in the task of legal article generation, enhancing the F1 score by 3\%. In terms of judgment, the retrieval of precedents can also assist the model in adjudicating cases. Furthermore, the analysis of the original case debates enables the model to better comprehend the cases, thereby improving the accuracy of the judgment outcomes. The evolution of the lawyer agent will elevate the quality of discourse, consequently augmenting the understanding of the cases.\\
\textbf{rare cases}. As illustrated in Figure \ref{fig3}, the rarer the data, the worse the model performs, which indicates that the model's capability to handle rare cases is insufficient and there is large room for improvement.\\
\textbf{Fine-tune Iteratively} 
We generated 1,000 cases with GPT-4o for argument. Initially, Qwen1.5-7B-Chat generated arguments for each case, creating 1,000 records. In a subsequent round, the fine-tuned Qwen1.5-7B-Chat generated further arguments for these same cases. The evolved agents' performance was then evaluated on 50 true cases, where we recorded the highest, lowest, and average argument scores.
As illustrated in Figure \ref{fig4}, as the tuning iterations progress, all three categories of scores have improved and gradually stabilized. However, in iteration 3, the assessment score declines, but is still better than the vanilla model.
In Table \ref{table5}, it is observed that the court arguments can enhance the Legal Judgment Prediction (LJP) ability of judge agents. Furthermore, GPT-4o demonstrates the capacity to generate more compelling arguments compared to smaller models such as Qwen1.5-7B-Chat.
Table \ref{table6} illustrates the impact of iterative evolution. While Iteration 2 does not outperform Iteration 1 across all tasks, both Iteration 1 and Iteration 2 demonstrate substantial improvements over the no-iteration baseline.

\begin{table*}[t]
\centering
\begin{small}
\begin{tabular}{lcccccccp{0.8cm}c}
\toprule
\textbf{Model} & \multicolumn{3}{c}{\textbf{Legal Articles}} & \multicolumn{3}{c}{\textbf{Civil and Admini.}} & \multicolumn{3}{c}{\textbf{Criminal}}  \\
 & P & R & F & P & R & F & Charge & Prison term & Fine \\
\midrule
Qwen1.5-7B-Chat & 0.41 & 0.12 & 0.18 & 0.26 & 0.36 & 0.28 & 0.83 & 0.06 & 0.05 \\
iteration1 & 0.41 & 0.12 & 0.17 & \textbf{0.33} & 0.38 & \textbf{0.34} & 0.90 & \textbf{0.10} & 0.06 \\
iteration2 & \textbf{0.41} & \textbf{0.13} & \textbf{0.20} & 0.30 & \textbf{0.40} & 0.32 & \textbf{0.95} & 0.06 & \textbf{0.06} \\
\midrule
\end{tabular}
\end{small} 
\caption{Performance of 120 Cases sampled from SimuCourt}
\label{table6}
\end{table*}

\begin{table*}[t]    
    \centering
    \begin{small}
    \begin{tabular}{l|c|cccc}
        \toprule
        Model & Legal Articles &  \multicolumn{3}{c}{Judgement Results} \\
        \cline{3-6}
        &  & Civil & Charge & term & Fine \\
        \midrule
        ASP2LJ & 0.127 & 0.167 & 0.850 & 0.161 & 0.107 \\
        w/o Court argument & 0.110 & 0.148 & 0.798 & 0.135 & 0.093 \\
        w/o Lawyer Evolution & 0.114 & 0.156 & 0.823 & 0.146 & 0.098 \\
        w/o Retriever & 0.094 & 0.125 & 0.779 & 0.130 & 0.055 \\
        \bottomrule
    \end{tabular}
    \end{small}
    \caption{Ablation Experiment on RareCases}
    \label{table7}
\end{table*}

\section{Conclusion}
We conduct a thorough analysis of our framework's performance. In our framework, lawyer agents can evolve and the judge can benefit from the evolution. To deal with the legal cases' long-tail distribution, we propose a method to gather legal cases by generating legal cases based on legal articles. Then We fine-tune the Qwen1.5-7B-Chat with the generated data to gain a better performance. The experimental results show that our method enables a weak model, Qwen1.5-7B-Chat, to surpass powerful models like GPT-4.  Besides, the proposed dataset, RareCases, also indicates that there is still an improvement room in the LJP tasks. 

\section{Limitations}
In this work, we primarily introduce an approach to generate cases automatically and propose a dataset encompassing rare cases. Despite our contribution, there is still some limitations. We just focus on Chinese laws while there are still various cases that are much different, leaving room to explore. We only focus on SimuCourt and our RareCases without evaluating other well-know datasets like LAiW\citep{dai2024laiwchineselegallarge}, LawBench or CAIL. Our framework incurs some time cost for the court argument process. We plan to explore the legal tasks in future studies further.

\bibliography{anthology,custom}

\begin{thebibliography}{45}
\expandafter\ifx\csname natexlab\endcsname\relax\def\natexlab#1{#1}\fi

\bibitem[{Anderson and Heaton(2012)}]{anderson2012much}
James~M Anderson and Paul Heaton. 2012.
\newblock How much difference does the lawyer make: The effect of defense counsel on murder case outcomes.
\newblock \emph{Yale LJ}, 122:154.

\bibitem[{Bai et~al.(2023)Bai, Bai, Chu, Cui, Dang, Deng, Fan, Ge, Han, Huang, Hui, Ji, Li, Lin, Lin, Liu, Liu, Lu, Lu, Ma, Men, Ren, Ren, Tan, Tan, Tu, Wang, Wang, Wang, Wu, Xu, Xu, Yang, Yang, Yang, Yang, Yao, Yu, Yuan, Yuan, Zhang, Zhang, Zhang, Zhang, Zhou, Zhou, Zhou, and Zhu}]{bai2023qwentechnicalreport}
Jinze Bai, Shuai Bai, Yunfei Chu, Zeyu Cui, Kai Dang, Xiaodong Deng, Yang Fan, Wenbin Ge, Yu~Han, Fei Huang, Binyuan Hui, Luo Ji, Mei Li, Junyang Lin, Runji Lin, Dayiheng Liu, Gao Liu, Chengqiang Lu, Keming Lu, Jianxin Ma, Rui Men, Xingzhang Ren, Xuancheng Ren, Chuanqi Tan, Sinan Tan, Jianhong Tu, Peng Wang, Shijie Wang, Wei Wang, Shengguang Wu, Benfeng Xu, Jin Xu, An~Yang, Hao Yang, Jian Yang, Shusheng Yang, Yang Yao, Bowen Yu, Hongyi Yuan, Zheng Yuan, Jianwei Zhang, Xingxuan Zhang, Yichang Zhang, Zhenru Zhang, Chang Zhou, Jingren Zhou, Xiaohuan Zhou, and Tianhang Zhu. 2023.
\newblock \href {http://arxiv.org/abs/2309.16609} {Qwen technical report}.

\bibitem[{B{\"u}ttner and Habernal(2024)}]{buttner-habernal-2024-answering}
Marius B{\"u}ttner and Ivan Habernal. 2024.
\newblock \href {https://aclanthology.org/2024.eacl-long.122/} {Answering legal questions from laymen in {G}erman civil law system}.
\newblock In \emph{Proceedings of the 18th Conference of the European Chapter of the Association for Computational Linguistics (Volume 1: Long Papers)}, pages 2015--2027, St. Julian{'}s, Malta. Association for Computational Linguistics.

\bibitem[{Chalkidis et~al.(2020)Chalkidis, Fergadiotis, Malakasiotis, Aletras, and Androutsopoulos}]{chalkidis2020legal}
Ilias Chalkidis, Manos Fergadiotis, Prodromos Malakasiotis, Nikolaos Aletras, and Ion Androutsopoulos. 2020.
\newblock Legal-bert: The muppets straight out of law school.
\newblock \emph{arXiv preprint arXiv:2010.02559}.

\bibitem[{Chen et~al.(2024{\natexlab{a}})Chen, Fan, Gong, Xie, Li, Liu, Li, Qu, Ni, and Yang}]{chen2024agentcourtsimulatingcourtadversarial}
Guhong Chen, Liyang Fan, Zihan Gong, Nan Xie, Zixuan Li, Ziqiang Liu, Chengming Li, Qiang Qu, Shiwen Ni, and Min Yang. 2024{\natexlab{a}}.
\newblock \href {http://arxiv.org/abs/2408.08089} {Agentcourt: Simulating court with adversarial evolvable lawyer agents}.

\bibitem[{Chen et~al.(2024{\natexlab{b}})Chen, Xiao, Zhang, Luo, Lian, and Liu}]{chen2024bgem3embeddingmultilingualmultifunctionality}
Jianlv Chen, Shitao Xiao, Peitian Zhang, Kun Luo, Defu Lian, and Zheng Liu. 2024{\natexlab{b}}.
\newblock \href {http://arxiv.org/abs/2402.03216} {Bge m3-embedding: Multi-lingual, multi-functionality, multi-granularity text embeddings through self-knowledge distillation}.

\bibitem[{Chen et~al.(2025)Chen, Mao, Li, and Shangguan}]{chen2025debatefeedbackmultiagentframeworkefficient}
Xi~Chen, Mao Mao, Shuo Li, and Haotian Shangguan. 2025.
\newblock \href {http://arxiv.org/abs/2504.05358} {Debate-feedback: A multi-agent framework for efficient legal judgment prediction}.

\bibitem[{Cui et~al.(2022)Cui, Shen, Nie, Wang, Wang, and Chen}]{cui2022surveylegaljudgmentprediction}
Junyun Cui, Xiaoyu Shen, Feiping Nie, Zheng Wang, Jinglong Wang, and Yulong Chen. 2022.
\newblock \href {http://arxiv.org/abs/2204.04859} {A survey on legal judgment prediction: Datasets, metrics, models and challenges}.

\bibitem[{Dai et~al.(2024)Dai, Feng, Huang, Jia, Xie, Zhang, Han, Tian, and Wang}]{dai2024laiwchineselegallarge}
Yongfu Dai, Duanyu Feng, Jimin Huang, Haochen Jia, Qianqian Xie, Yifang Zhang, Weiguang Han, Wei Tian, and Hao Wang. 2024.
\newblock \href {http://arxiv.org/abs/2310.05620} {Laiw: A chinese legal large language models benchmark}.

\bibitem[{Douka et~al.(2021)Douka, Abdine, Vazirgiannis, Hamdani, and Amariles}]{douka2021juribert}
Stella Douka, Hadi Abdine, Michalis Vazirgiannis, Rajaa~El Hamdani, and David~Restrepo Amariles. 2021.
\newblock Juribert: A masked-language model adaptation for french legal text.
\newblock \emph{arXiv preprint arXiv:2110.01485}.

\bibitem[{Fei et~al.(2023)Fei, Shen, Zhu, Zhou, Han, Zhang, Chen, Shen, and Ge}]{fei2023lawbenchbenchmarkinglegalknowledge}
Zhiwei Fei, Xiaoyu Shen, Dawei Zhu, Fengzhe Zhou, Zhuo Han, Songyang Zhang, Kai Chen, Zongwen Shen, and Jidong Ge. 2023.
\newblock \href {http://arxiv.org/abs/2309.16289} {Lawbench: Benchmarking legal knowledge of large language models}.

\bibitem[{Feng et~al.(2022)Feng, Li, and Ng}]{feng2022legal}
Yi~Feng, Chuanyi Li, and Vincent Ng. 2022.
\newblock Legal judgment prediction via event extraction with constraints.
\newblock In \emph{Proceedings of the 60th annual meeting of the association for computational linguistics (volume 1: long papers)}, pages 648--664.

\bibitem[{Feng et~al.(2024)Feng, Li, and Ng}]{feng-etal-2024-legal}
Yi~Feng, Chuanyi Li, and Vincent Ng. 2024.
\newblock \href {https://doi.org/10.18653/v1/2024.acl-long.350} {Legal case retrieval: A survey of the state of the art}.
\newblock In \emph{Proceedings of the 62nd Annual Meeting of the Association for Computational Linguistics (Volume 1: Long Papers)}, pages 6472--6485, Bangkok, Thailand. Association for Computational Linguistics.

\bibitem[{Gao et~al.(2024)Gao, Xiao, Liu, Chen, Liu, and Sun}]{gao2024enhancinglegalcaseretrieval}
Cheng Gao, Chaojun Xiao, Zhenghao Liu, Huimin Chen, Zhiyuan Liu, and Maosong Sun. 2024.
\newblock \href {http://arxiv.org/abs/2410.06581} {Enhancing legal case retrieval via scaling high-quality synthetic query-candidate pairs}.

\bibitem[{Garrett(2011)}]{garrett2011convicting}
Brandon~L Garrett. 2011.
\newblock \emph{Convicting the innocent: Where criminal prosecutions go wrong}.
\newblock Harvard University Press.

\bibitem[{Gross et~al.(2014)Gross, O’brien, Hu, and Kennedy}]{gross2014rate}
Samuel~R Gross, Barbara O’brien, Chen Hu, and Edward~H Kennedy. 2014.
\newblock Rate of false conviction of criminal defendants who are sentenced to death.
\newblock \emph{Proceedings of the National Academy of Sciences}, 111(20):7230--7235.

\bibitem[{Habernal et~al.(2024)Habernal, Faber, Recchia, Bretthauer, Gurevych, Spiecker~genannt D{\"o}hmann, and Burchard}]{habernal2024mining}
Ivan Habernal, Daniel Faber, Nicola Recchia, Sebastian Bretthauer, Iryna Gurevych, Indra Spiecker~genannt D{\"o}hmann, and Christoph Burchard. 2024.
\newblock Mining legal arguments in court decisions.
\newblock \emph{Artificial Intelligence and Law}, 32(3):1--38.

\bibitem[{He et~al.(2024)He, Cao, Wang, Jin, Chen, Xu, Li, Liu, and Zhao}]{he-etal-2024-agentscourt}
Zhitao He, Pengfei Cao, Chenhao Wang, Zhuoran Jin, Yubo Chen, Jiexin Xu, Huaijun Li, Kang Liu, and Jun Zhao. 2024.
\newblock \href {https://doi.org/10.18653/v1/2024.findings-emnlp.549} {{A}gents{C}ourt: Building judicial decision-making agents with court debate simulation and legal knowledge augmentation}.
\newblock In \emph{Findings of the Association for Computational Linguistics: EMNLP 2024}, pages 9399--9416, Miami, Florida, USA. Association for Computational Linguistics.

\bibitem[{Hou et~al.(2024)Hou, Weller, Qin, Yang, Lawrie, Holzenberger, Blair-Stanek, and Durme}]{hou2024clercdatasetlegalcase}
Abe~Bohan Hou, Orion Weller, Guanghui Qin, Eugene Yang, Dawn Lawrie, Nils Holzenberger, Andrew Blair-Stanek, and Benjamin~Van Durme. 2024.
\newblock \href {http://arxiv.org/abs/2406.17186} {Clerc: A dataset for legal case retrieval and retrieval-augmented analysis generation}.

\bibitem[{Huang et~al.(2024)Huang, Feng, Li, Wu, Ge, and Ng}]{huang-etal-2024-cmdl}
Wanhong Huang, Yi~Feng, Chuanyi Li, Honghan Wu, Jidong Ge, and Vincent Ng. 2024.
\newblock \href {https://doi.org/10.18653/v1/2024.findings-acl.351} {{CMDL}: A large-scale {C}hinese multi-defendant legal judgment prediction dataset}.
\newblock In \emph{Findings of the Association for Computational Linguistics: ACL 2024}, pages 5895--5906, Bangkok, Thailand. Association for Computational Linguistics.

\bibitem[{Kim et~al.(2024)Kim, Choi, Choi, Choi, Park, and Hwang}]{kim2024developingpragmaticbenchmarkassessing}
Yeeun Kim, Young~Rok Choi, Eunkyung Choi, Jinhwan Choi, Hai~Jin Park, and Wonseok Hwang. 2024.
\newblock \href {http://arxiv.org/abs/2410.08731} {Developing a pragmatic benchmark for assessing korean legal language understanding in large language models}.

\bibitem[{Li et~al.(2025{\natexlab{a}})Li, Wu, Cai, Jatowt, Zhou, Lu, Sun, Wu, and Kuang}]{li-etal-2025-legal}
Ang Li, Yiquan Wu, Ming Cai, Adam Jatowt, Xiang Zhou, Weiming Lu, Changlong Sun, Fei Wu, and Kun Kuang. 2025{\natexlab{a}}.
\newblock \href {https://aclanthology.org/2025.naacl-long.355/} {Legal judgment prediction based on knowledge-enhanced multi-task and multi-label text classification}.
\newblock In \emph{Proceedings of the 2025 Conference of the Nations of the Americas Chapter of the Association for Computational Linguistics: Human Language Technologies (Volume 1: Long Papers)}, pages 6957--6970, Albuquerque, New Mexico. Association for Computational Linguistics.

\bibitem[{Li et~al.(2023)Li, Shao, Wu, Ai, Ma, and Liu}]{li2023lecardv2largescalechineselegal}
Haitao Li, Yunqiu Shao, Yueyue Wu, Qingyao Ai, Yixiao Ma, and Yiqun Liu. 2023.
\newblock \href {http://arxiv.org/abs/2310.17609} {Lecardv2: A large-scale chinese legal case retrieval dataset}.

\bibitem[{Li et~al.(2025{\natexlab{b}})Li, Mi, Meng, Jia, Zhao, Zhao, Wei, Gao, and Li}]{Li2025AddressingLD}
Ting Li, Lewen Mi, Xiangyu Meng, Yongju Jia, Lin Zhao, Qi~Zhao, Zihao Wei, Guandong Gao, and Xiangxian Li. 2025{\natexlab{b}}.
\newblock \href {https://api.semanticscholar.org/CorpusID:276486735} {Addressing long-tailed distribution in judicial text for criminal motive classification: a balanced contrastive learning approach}.
\newblock \emph{EPJ Data Sci.}, 14:14.

\bibitem[{{Li, Ting} et~al.(2025){Li, Ting}, {Mi, Lewen}, {Meng, Xiangyu}, {Jia, Yongju}, {Zhao, Lin}, {Zhao, Qi}, {Wei, Zihao}, {Gao, Guandong}, and {Li, Xiangxian}}]{refId0}
{Li, Ting}, {Mi, Lewen}, {Meng, Xiangyu}, {Jia, Yongju}, {Zhao, Lin}, {Zhao, Qi}, {Wei, Zihao}, {Gao, Guandong}, and {Li, Xiangxian}. 2025.
\newblock \href {https://doi.org/10.1140/epjds/s13688-025-00533-1} {Addressing long-tailed distribution in judicial text for criminal motive classification: a balanced contrastive learning approach}.
\newblock \emph{EPJ Data Sci.}, 14(1):14.

\bibitem[{Limsopatham(2021)}]{limsopatham2021effectively}
Nut Limsopatham. 2021.
\newblock Effectively leveraging bert for legal document classification.
\newblock In \emph{Proceedings of the Natural Legal Language Processing Workshop 2021}, pages 210--216.

\bibitem[{Ma et~al.(2021)Ma, Zhang, Wang, Liu, Ye, Sun, and Zhang}]{Ma_2021}
Luyao Ma, Yating Zhang, Tianyi Wang, Xiaozhong Liu, Wei Ye, Changlong Sun, and Shikun Zhang. 2021.
\newblock \href {https://doi.org/10.1145/3404835.3462945} {Legal judgment prediction with multi-stage case representation learning in the real court setting}.
\newblock In \emph{Proceedings of the 44th International ACM SIGIR Conference on Research and Development in Information Retrieval}, SIGIR ’21, page 993–1002. ACM.

\bibitem[{Mistica et~al.(2020)Mistica, Zhang, Chia, Shrestha, Gupta, Khandelwal, Paterson, Baldwin, and Beck}]{mistica2020information}
Meladel Mistica, Geordie~Z Zhang, Hui Chia, Kabir~Manandhar Shrestha, Rohit~Kumar Gupta, Saket Khandelwal, Jeannie Paterson, Timothy Baldwin, and Daniel Beck. 2020.
\newblock Information extraction from legal documents: A study in the context of common law court judgements.
\newblock In \emph{Proceedings of the 18th annual workshop of the australasian language technology association}, pages 98--103.

\bibitem[{Niklaus et~al.(2021)Niklaus, Chalkidis, and Stürmer}]{niklaus2021swissjudgmentpredictionmultilinguallegaljudgment}
Joel Niklaus, Ilias Chalkidis, and Matthias Stürmer. 2021.
\newblock \href {http://arxiv.org/abs/2110.00806} {Swiss-judgment-prediction: A multilingual legal judgment prediction benchmark}.

\bibitem[{OpenAI et~al.(2024)OpenAI, Achiam, Adler, Agarwal, Ahmad, Akkaya, Aleman, Almeida, Altenschmidt, Altman, Anadkat, Avila, Babuschkin, Balaji, Balcom, Baltescu, Bao, Bavarian, Belgum, Bello, Berdine, Bernadett-Shapiro, Berner, Bogdonoff, Boiko, Boyd, Brakman, Brockman, Brooks, Brundage, Button, Cai, Campbell, Cann, Carey, Carlson, Carmichael, Chan, Chang, Chantzis, Chen, Chen, Chen, Chen, Chen, Chess, Cho, Chu, Chung, Cummings, Currier, Dai, Decareaux, Degry, Deutsch, Deville, Dhar, Dohan, Dowling, Dunning, Ecoffet, Eleti, Eloundou, Farhi, Fedus, Felix, Fishman, Forte, Fulford, Gao, Georges, Gibson, Goel, Gogineni, Goh, Gontijo-Lopes, Gordon, Grafstein, Gray, Greene, Gross, Gu, Guo, Hallacy, Han, Harris, He, Heaton, Heidecke, Hesse, Hickey, Hickey, Hoeschele, Houghton, Hsu, Hu, Hu, Huizinga, Jain, Jain, Jang, Jiang, Jiang, Jin, Jin, Jomoto, Jonn, Jun, Kaftan, Łukasz Kaiser, Kamali, Kanitscheider, Keskar, Khan, Kilpatrick, Kim, Kim, Kim, Kirchner, Kiros, Knight, Kokotajlo, Łukasz Kondraciuk,
  Kondrich, Konstantinidis, Kosic, Krueger, Kuo, Lampe, Lan, Lee, Leike, Leung, Levy, Li, Lim, Lin, Lin, Litwin, Lopez, Lowe, Lue, Makanju, Malfacini, Manning, Markov, Markovski, Martin, Mayer, Mayne, McGrew, McKinney, McLeavey, McMillan, McNeil, Medina, Mehta, Menick, Metz, Mishchenko, Mishkin, Monaco, Morikawa, Mossing, Mu, Murati, Murk, Mély, Nair, Nakano, Nayak, Neelakantan, Ngo, Noh, Ouyang, O'Keefe, Pachocki, Paino, Palermo, Pantuliano, Parascandolo, Parish, Parparita, Passos, Pavlov, Peng, Perelman, de~Avila Belbute~Peres, Petrov, de~Oliveira~Pinto, Michael, Pokorny, Pokrass, Pong, Powell, Power, Power, Proehl, Puri, Radford, Rae, Ramesh, Raymond, Real, Rimbach, Ross, Rotsted, Roussez, Ryder, Saltarelli, Sanders, Santurkar, Sastry, Schmidt, Schnurr, Schulman, Selsam, Sheppard, Sherbakov, Shieh, Shoker, Shyam, Sidor, Sigler, Simens, Sitkin, Slama, Sohl, Sokolowsky, Song, Staudacher, Such, Summers, Sutskever, Tang, Tezak, Thompson, Tillet, Tootoonchian, Tseng, Tuggle, Turley, Tworek, Uribe, Vallone,
  Vijayvergiya, Voss, Wainwright, Wang, Wang, Wang, Ward, Wei, Weinmann, Welihinda, Welinder, Weng, Weng, Wiethoff, Willner, Winter, Wolrich, Wong, Workman, Wu, Wu, Wu, Xiao, Xu, Yoo, Yu, Yuan, Zaremba, Zellers, Zhang, Zhang, Zhao, Zheng, Zhuang, Zhuk, and Zoph}]{openai2024gpt4technicalreport}
OpenAI, Josh Achiam, Steven Adler, Sandhini Agarwal, Lama Ahmad, Ilge Akkaya, Florencia~Leoni Aleman, Diogo Almeida, Janko Altenschmidt, Sam Altman, Shyamal Anadkat, Red Avila, Igor Babuschkin, Suchir Balaji, Valerie Balcom, Paul Baltescu, Haiming Bao, Mohammad Bavarian, Jeff Belgum, Irwan Bello, Jake Berdine, Gabriel Bernadett-Shapiro, Christopher Berner, Lenny Bogdonoff, Oleg Boiko, Madelaine Boyd, Anna-Luisa Brakman, Greg Brockman, Tim Brooks, Miles Brundage, Kevin Button, Trevor Cai, Rosie Campbell, Andrew Cann, Brittany Carey, Chelsea Carlson, Rory Carmichael, Brooke Chan, Che Chang, Fotis Chantzis, Derek Chen, Sully Chen, Ruby Chen, Jason Chen, Mark Chen, Ben Chess, Chester Cho, Casey Chu, Hyung~Won Chung, Dave Cummings, Jeremiah Currier, Yunxing Dai, Cory Decareaux, Thomas Degry, Noah Deutsch, Damien Deville, Arka Dhar, David Dohan, Steve Dowling, Sheila Dunning, Adrien Ecoffet, Atty Eleti, Tyna Eloundou, David Farhi, Liam Fedus, Niko Felix, Simón~Posada Fishman, Juston Forte, Isabella Fulford, Leo
  Gao, Elie Georges, Christian Gibson, Vik Goel, Tarun Gogineni, Gabriel Goh, Rapha Gontijo-Lopes, Jonathan Gordon, Morgan Grafstein, Scott Gray, Ryan Greene, Joshua Gross, Shixiang~Shane Gu, Yufei Guo, Chris Hallacy, Jesse Han, Jeff Harris, Yuchen He, Mike Heaton, Johannes Heidecke, Chris Hesse, Alan Hickey, Wade Hickey, Peter Hoeschele, Brandon Houghton, Kenny Hsu, Shengli Hu, Xin Hu, Joost Huizinga, Shantanu Jain, Shawn Jain, Joanne Jang, Angela Jiang, Roger Jiang, Haozhun Jin, Denny Jin, Shino Jomoto, Billie Jonn, Heewoo Jun, Tomer Kaftan, Łukasz Kaiser, Ali Kamali, Ingmar Kanitscheider, Nitish~Shirish Keskar, Tabarak Khan, Logan Kilpatrick, Jong~Wook Kim, Christina Kim, Yongjik Kim, Jan~Hendrik Kirchner, Jamie Kiros, Matt Knight, Daniel Kokotajlo, Łukasz Kondraciuk, Andrew Kondrich, Aris Konstantinidis, Kyle Kosic, Gretchen Krueger, Vishal Kuo, Michael Lampe, Ikai Lan, Teddy Lee, Jan Leike, Jade Leung, Daniel Levy, Chak~Ming Li, Rachel Lim, Molly Lin, Stephanie Lin, Mateusz Litwin, Theresa Lopez, Ryan
  Lowe, Patricia Lue, Anna Makanju, Kim Malfacini, Sam Manning, Todor Markov, Yaniv Markovski, Bianca Martin, Katie Mayer, Andrew Mayne, Bob McGrew, Scott~Mayer McKinney, Christine McLeavey, Paul McMillan, Jake McNeil, David Medina, Aalok Mehta, Jacob Menick, Luke Metz, Andrey Mishchenko, Pamela Mishkin, Vinnie Monaco, Evan Morikawa, Daniel Mossing, Tong Mu, Mira Murati, Oleg Murk, David Mély, Ashvin Nair, Reiichiro Nakano, Rajeev Nayak, Arvind Neelakantan, Richard Ngo, Hyeonwoo Noh, Long Ouyang, Cullen O'Keefe, Jakub Pachocki, Alex Paino, Joe Palermo, Ashley Pantuliano, Giambattista Parascandolo, Joel Parish, Emy Parparita, Alex Passos, Mikhail Pavlov, Andrew Peng, Adam Perelman, Filipe de~Avila Belbute~Peres, Michael Petrov, Henrique~Ponde de~Oliveira~Pinto, Michael, Pokorny, Michelle Pokrass, Vitchyr~H. Pong, Tolly Powell, Alethea Power, Boris Power, Elizabeth Proehl, Raul Puri, Alec Radford, Jack Rae, Aditya Ramesh, Cameron Raymond, Francis Real, Kendra Rimbach, Carl Ross, Bob Rotsted, Henri Roussez,
  Nick Ryder, Mario Saltarelli, Ted Sanders, Shibani Santurkar, Girish Sastry, Heather Schmidt, David Schnurr, John Schulman, Daniel Selsam, Kyla Sheppard, Toki Sherbakov, Jessica Shieh, Sarah Shoker, Pranav Shyam, Szymon Sidor, Eric Sigler, Maddie Simens, Jordan Sitkin, Katarina Slama, Ian Sohl, Benjamin Sokolowsky, Yang Song, Natalie Staudacher, Felipe~Petroski Such, Natalie Summers, Ilya Sutskever, Jie Tang, Nikolas Tezak, Madeleine~B. Thompson, Phil Tillet, Amin Tootoonchian, Elizabeth Tseng, Preston Tuggle, Nick Turley, Jerry Tworek, Juan Felipe~Cerón Uribe, Andrea Vallone, Arun Vijayvergiya, Chelsea Voss, Carroll Wainwright, Justin~Jay Wang, Alvin Wang, Ben Wang, Jonathan Ward, Jason Wei, CJ~Weinmann, Akila Welihinda, Peter Welinder, Jiayi Weng, Lilian Weng, Matt Wiethoff, Dave Willner, Clemens Winter, Samuel Wolrich, Hannah Wong, Lauren Workman, Sherwin Wu, Jeff Wu, Michael Wu, Kai Xiao, Tao Xu, Sarah Yoo, Kevin Yu, Qiming Yuan, Wojciech Zaremba, Rowan Zellers, Chong Zhang, Marvin Zhang, Shengjia
  Zhao, Tianhao Zheng, Juntang Zhuang, William Zhuk, and Barret Zoph. 2024.
\newblock \href {http://arxiv.org/abs/2303.08774} {Gpt-4 technical report}.

\bibitem[{Pipitone and Alami(2024)}]{pipitone2024legalbenchragbenchmarkretrievalaugmentedgeneration}
Nicholas Pipitone and Ghita~Houir Alami. 2024.
\newblock \href {http://arxiv.org/abs/2408.10343} {Legalbench-rag: A benchmark for retrieval-augmented generation in the legal domain}.

\bibitem[{Poppe and Rachlinski(2015)}]{poppe2015lawyers}
Emily S~Taylor Poppe and Jeffrey~J Rachlinski. 2015.
\newblock Do lawyers matter? the effect of legal representation in civil disputes.
\newblock \emph{Pepp. L. Rev.}, 43:881.

\bibitem[{Qin et~al.(2024)Qin, Cao, Yu, Si, Chen, and Xu}]{Qin_2024}
Weicong Qin, Zelin Cao, Weijie Yu, Zihua Si, Sirui Chen, and Jun Xu. 2024.
\newblock \href {https://doi.org/10.1145/3626772.3657717} {Explicitly integrating judgment prediction with legal document retrieval: A law-guided generative approach}.
\newblock In \emph{Proceedings of the 47th International ACM SIGIR Conference on Research and Development in Information Retrieval}, SIGIR 2024, page 2210–2220. ACM.

\bibitem[{Rosa et~al.(2021)Rosa, Rodrigues, Lotufo, and Nogueira}]{rosa2021yesbm25strongbaseline}
Guilherme~Moraes Rosa, Ruan~Chaves Rodrigues, Roberto Lotufo, and Rodrigo Nogueira. 2021.
\newblock \href {http://arxiv.org/abs/2105.05686} {Yes, bm25 is a strong baseline for legal case retrieval}.

\bibitem[{Semo et~al.(2022)Semo, Bernsohn, Hagag, Hayat, and Niklaus}]{semo-etal-2022-classactionprediction}
Gil Semo, Dor Bernsohn, Ben Hagag, Gila Hayat, and Joel Niklaus. 2022.
\newblock \href {https://doi.org/10.18653/v1/2022.nllp-1.3} {{C}lass{A}ction{P}rediction: A challenging benchmark for legal judgment prediction of class action cases in the {US}}.
\newblock In \emph{Proceedings of the Natural Legal Language Processing Workshop 2022}, pages 31--46, Abu Dhabi, United Arab Emirates (Hybrid). Association for Computational Linguistics.

\bibitem[{Sheppard and Moshirnia(2012)}]{sheppard2012sake}
Brian Sheppard and Andrew Moshirnia. 2012.
\newblock For the sake of argument: A behavioral analysis of whether and how legal argument matters in decisionmaking.
\newblock \emph{Fla. St. UL Rev}, 40:537.

\bibitem[{Shiu-fan(1983)}]{shiu1983role}
Jenkin~Chan Shiu-fan. 1983.
\newblock The role of lawyers in the chinese legal system.
\newblock \emph{Hong Kong LJ}, 13:157.

\bibitem[{Wang et~al.(2024)Wang, Su, Yeh, and Fan}]{wang2024crosslingualstatutoryarticleretrieval}
Yen-Hsiang Wang, Feng-Dian Su, Tzu-Yu Yeh, and Yao-Chung Fan. 2024.
\newblock \href {http://arxiv.org/abs/2410.11450} {A cross-lingual statutory article retrieval dataset for taiwan legal studies}.

\bibitem[{Wu et~al.(2023)Wu, Zhou, Liu, Lu, Liu, Zhang, Sun, Wu, and Kuang}]{wu2023precedentenhancedlegaljudgmentprediction}
Yiquan Wu, Siying Zhou, Yifei Liu, Weiming Lu, Xiaozhong Liu, Yating Zhang, Changlong Sun, Fei Wu, and Kun Kuang. 2023.
\newblock \href {http://arxiv.org/abs/2310.09241} {Precedent-enhanced legal judgment prediction with llm and domain-model collaboration}.

\bibitem[{Xiao et~al.(2018)Xiao, Zhong, Guo, Tu, Liu, Sun, Feng, Han, Hu, Wang, and Xu}]{xiao2018cail2018largescalelegaldataset}
Chaojun Xiao, Haoxi Zhong, Zhipeng Guo, Cunchao Tu, Zhiyuan Liu, Maosong Sun, Yansong Feng, Xianpei Han, Zhen Hu, Heng Wang, and Jianfeng Xu. 2018.
\newblock \href {http://arxiv.org/abs/1807.02478} {Cail2018: A large-scale legal dataset for judgment prediction}.

\bibitem[{Xu et~al.(2020)Xu, Wang, Chen, Pan, Wang, and Zhao}]{xu2020distinguish}
Nuo Xu, Pinghui Wang, Long Chen, Li~Pan, Xiaoyan Wang, and Junzhou Zhao. 2020.
\newblock Distinguish confusing law articles for legal judgment prediction.
\newblock \emph{arXiv preprint arXiv:2004.02557}.

\bibitem[{Yao et~al.(2023)Yao, Zhao, Yu, Du, Shafran, Narasimhan, and Cao}]{yao2023reactsynergizingreasoningacting}
Shunyu Yao, Jeffrey Zhao, Dian Yu, Nan Du, Izhak Shafran, Karthik Narasimhan, and Yuan Cao. 2023.
\newblock \href {http://arxiv.org/abs/2210.03629} {React: Synergizing reasoning and acting in language models}.

\bibitem[{Zavr{\v{s}}nik(2021)}]{zavrvsnik2021algorithmic}
Ale{\v{s}} Zavr{\v{s}}nik. 2021.
\newblock Algorithmic justice: Algorithms and big data in criminal justice settings.
\newblock \emph{European Journal of criminology}, 18(5):623--642.

\bibitem[{Zhong et~al.(2018)Zhong, Guo, Tu, Xiao, Liu, and Sun}]{zhong2018legal}
Haoxi Zhong, Zhipeng Guo, Cunchao Tu, Chaojun Xiao, Zhiyuan Liu, and Maosong Sun. 2018.
\newblock Legal judgment prediction via topological learning.
\newblock In \emph{Proceedings of the 2018 conference on empirical methods in natural language processing}, pages 3540--3549.

\bibitem[{Zhou et~al.(2024)Zhou, Shi, Song, Yang, Jin, Guo, and Li}]{zhou2024lawgptchineselegalknowledgeenhanced}
Zhi Zhou, Jiang-Xin Shi, Peng-Xiao Song, Xiao-Wen Yang, Yi-Xuan Jin, Lan-Zhe Guo, and Yu-Feng Li. 2024.
\newblock \href {http://arxiv.org/abs/2406.04614} {Lawgpt: A chinese legal knowledge-enhanced large language model}.

\end{thebibliography}
\bibliographystyle{acl_natbib}

\section{Related Work}
Legal Artificial Intelligence is a rapidly growing field that has gathered significant interest among researchers, encompassing various tasks such as legal case retrieval (LCR), statutory article retrieval (SAR), and legal judgment prediction (LJP). 
\subsection{Legal AI}
Prior to the advent of LLMs, legal tasks were predominantly addressed using conventional artificial intelligence techniques. 
CAIL\citep{xiao2018cail2018largescalelegaldataset} was established as a well-known annual Chinese legal AI competition, featuring tasks like LCR and LJP, which attracts widespread participation from legal AI researchers.
Some studies \citep{niklaus2021swissjudgmentpredictionmultilinguallegaljudgment} focus on the legal language varies in different countries, trying to construct benchmarks to evaluate the concurrent models' capacity in dealing with different language. Some works \citep{chalkidis2020legal, douka2021juribert, limsopatham2021effectively}  try to introduce specific retriever models like Bert into legal tasks
\subsection{LLM+Law}
Following the introduction of ChatGPT, numerous studies have explored integrating LLMs into legal tasks, yielding promising results. Due to LLM's strong performance on reasoning, \citet{yao2023reactsynergizingreasoningacting} combines LLM with legal knowledge. For instance, LawBench\citep{fei2023lawbenchbenchmarkinglegalknowledge} comprises approximately 20 tasks focused on legal memory, understanding, and application. GEAR\citep{Qin_2024} introduces a methodology that constructs a hierarchical structure of legal articles, thereby augmenting the model's interpretative capabilities. \citet{zhou2024lawgptchineselegalknowledgeenhanced} proposes LawGPT by fine-tuning Chinese-LLaMA with Chinese legal knowledge.
Additionally, several works \citep{li2023lecardv2largescalechineselegal, pipitone2024legalbenchragbenchmarkretrievalaugmentedgeneration, feng-etal-2024-legal, hou2024clercdatasetlegalcase, gao2024enhancinglegalcaseretrieval} have contributed to enhancing the retrieval capabilities of legal systems. LLMs can further improve their performance through retrieval-augmented generation (RAG). Concurrently, some researchers have explored using LLMs to tackle legal entrance exam questions\citep{kim2024developingpragmaticbenchmarkassessing}, but the performance is not satisfying, indicating the large challenge in the legal field. Other studies, such as \citet{wu2023precedentenhancedlegaljudgmentprediction}, have demonstrated that combining LLMs with domain-specific legal models can enhance LJP performance, while \citet{Qin_2024} introduced GEAR, a novel framework integrating LCR, SAR, and LJP.

\subsection{Legal Agent}
With the advent of LLM-based agents, researchers try to simulate courtroom environments using these agents. For example, \citet{chen2024agentcourtsimulatingcourtadversarial} employs agents to engage in debates and generate extensive records to refine their capabilities. Similarly, \citet{he-etal-2024-agentscourt} proposes a framework where lawyer agents argue, and the judge retrieves relevant legal articles, precedents, and papers to ensure the accuracy of the final judgment. Recently, \citet{chen2025debatefeedbackmultiagentframeworkefficient} focus on multi-agent debate to improve the judges' judgment precision by providing different views and lawyer agents' self-feedback. However, these works overlook the critical role of lawyers which results in suboptimal performance, leaving room for further improvement.

\appendix
\section{Evaluation Prompt}
\label{sec:appendix A}

A scoring criterion encompassing citation, refutation, and comprehension is proposed for a comprehensive and objective evaluation of lawyer agents' discourse. Specifically, this criterion assesses: \\
\textbf{Citation}: The inclusion of relevant articles or precedents. \\
\textbf{Refutation}: The act of rebutting the opposing counsel's arguments. \\
\textbf{Comprehension}: The provision of a comprehensive and clarified case description.

As shown in Table \ref{table5}, \ref{table6}, and \ref{table7}, our results demonstrate that the evolved lawyer agents can produce more effective arguments and improve lawyer agents' performance in LJP tasks.
We present our scoring criterion in Table \ref{table10}.

\section{Generated Case}
\label{sec:appendix B}

To verify our generated cases' rationality, we invite three undergraduate students major in law to evaluate the cases' quality. Our criterion is in three dimensions:  \\
\textbf{Correctness}: Make true if the case can happen in the real world. \\
\textbf{Reality}: Make true if there are relevant laws as a basis. \\
\textbf{Rationality}: Make true if the plaintiff and defendant's claims are within a reasonable legal framework. \\
We randomly sample 500 generated cases, and the evaluation results are presented in Table \ref{table8}.

\section{Data Analysis}
\label{sec:appendix C}

We collect all Chinese cases from the China Judgements Online and conduct a statistical analysis of the frequency of the case causes from 2018 to 2021. The least frequently occurring causes are identified and categorized as rare case causes. We collect 166 cases entailing rare causes from the WenShu web which occurred after 2022. Then we invite three undergraduate students to help us check and filter some cases and instruct GPT-4o to transform the cases into specific structure. Finally we get 120 cases, including 90 civil cases and 30 criminal cases.

\section{Metric}
\label{sec:appendix D}
We have three LJP tasks: law articles, final judgment, and case analysis. \\
\textbf{legal articles}. We use regular expressions to extract law articles from generated law articles and check whether the extracted articles match our answers or not. Then we calculate the answers' precision, recall, and F1.
\begin{equation}
\mathrm{F1} = 
\frac{
    2 \times \mathit{precision} \times \mathit{recall}
}{
    \mathit{precision} + \mathit{recall}
}
\label{eq:f1}
\end{equation}

\textbf{judgment results}. For criminal cases, the results include charge, prison term, and fine. We utilize regular expressions to extract these items from generated answer, and calculate the accuracy.
\begin{equation}
\mathrm{ACC} = 
\frac{
    \begin{tabular}{@{}c@{}}
    Number of correct answers
    \end{tabular}
}{
    \begin{tabular}{@{}c@{}}
    Total number of cases
    \end{tabular}
} \label{eq:f1}
\end{equation}

For civil and administrative cases, the answers are flexible and hard to extract by regular expressions. So we employ GPT-4o to summarize the answers as several points and judge the number of correct answers. Detailed prompts are presented below.

\begin{tcolorbox}[colback=gray!10!white,colframe=gray!80!black,title=Prompt 1]
Please organize the given text into the required format. \\
Example 1: Current text: The judgment is as follows: Defendant should return the loan of 200000 yuan to Plaintiff; Defendant shall pay interest during the period of fund occupation at an annual rate of 6\% from December 20, 2021 to October 19, 2023; The defendant shall bear all the litigation costs of this case. The above is the final judgment of this court. The defendant is requested to fulfill the repayment obligation within the time limit given in the judgment and pay interest and litigation costs in accordance with the law. \\
Output list: {"Result 1": "The defendant should return the loan of 200000 yuan to the plaintiff", "Result 2": "The defendant  should pay interest on the funds during the occupation period at an annual interest rate of 6\% from December 20, 2021 to October 19, 2023"} \\
Example 2: \dots \\
Example 3: \dots \\
Please organize the following content: \\
Current text:<RAW-RELUSTS> \\
Output List:
\end{tcolorbox}

\begin{tcolorbox}[colback=gray!10!white,colframe=gray!80!black,title=Prompt 2]
Please compare the candidate's answer with the reference answer to determine if the answer is correct. No explanation is needed, and the result can be directly output in JSON structure \\ 
Example 1:
Current text: Reference answers: {"Result 1": "The defendant should return the loan of 200000 yuan to the plaintiff", "Result 2": "The defendant should pay interest on the capital occupation period at an annual interest rate of 6\% from December 20, 2021 to October 19, 2023"} \\
Candidate answers: {"Result 1": "The defendant should pay interest on the capital occupation period at an annual interest rate of 6\% from December 20, 2021 to October 19, 2023", "Result 2": "The defendant should return the loan of 10000 yuan to the plaintiff"} \\
Output list: {Result 1: 0, Result 2: 1} \\
Example 2: ... \\
Example 3: ...\\
Please organize the following content and output it in JSON structure: \\
Current text:<RAW-RELUSTS> \\
Output list: {"Result 1":<>, "Result 2":<>,...}
\end{tcolorbox}

\textbf{case analysis} To verify agents' comprehension of the legal cases, we instruct agents to generate case analysis and invite three law school undergraduate students to assess. The scoring criterion is the same as AgentsCourt \citep{he-etal-2024-agentscourt}: 
1) Correctness: Mark true if and only if the analysis is satisfying and considers all parties involved. 2) Logicality: Mark false if the analysis contains any illogical or untrue reasoning. 3) Concision: Mark true if the analysis covers all necessary information without any extra information.

\section{Case Distribution}
\label{sec:appendix F}

\begin{figure*}[htbp]
	\centering  
	\includegraphics[width=17cm, height=10cm]{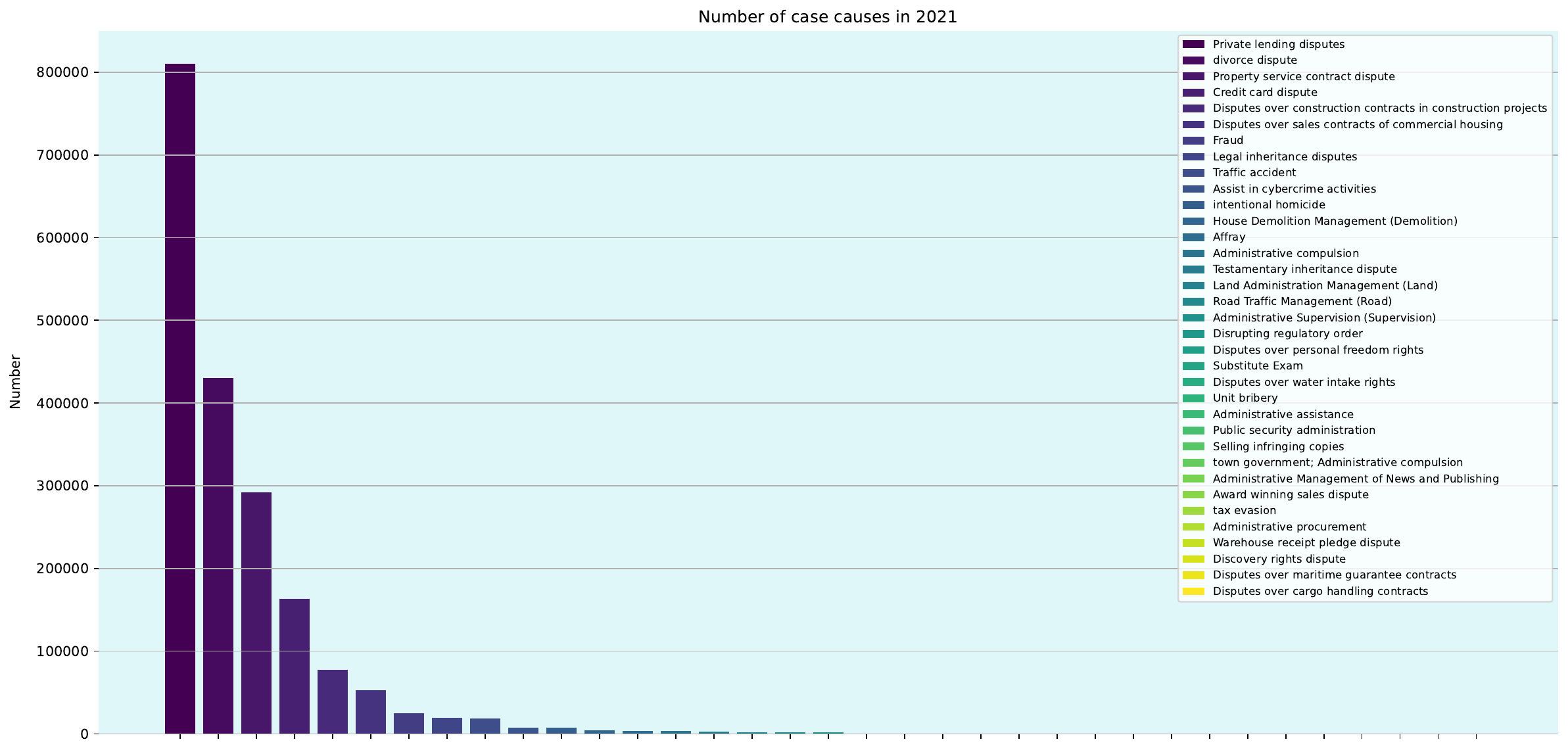}  
	\caption{Number of sampled cases in 2021.}
	\label{longtail}
\end{figure*}

As illustrated in Figure \ref{longtail}, the case distribution exhibits a long-tail characteristic. For instance, 'Legal inheritance disputes' accounts for 809,843 instances, whereas 'Disputes over cargo handling contracts' comprises only one.

\section{Argument Evaluation Score}
\label{sec:appendix F}
We present our evaluation score in Table \ref{table10}.

\begin{table*}[h]
\centering
\begin{normalsize}
\begin{tabular}{lp{0.9\textwidth}}
\toprule
\textbf{Model} \\
\midrule
s1:  Did the content of the debate cite legal provisions and cases, and is the relevance of legal articles \\
and cases to this case significant? Can you provide effective assistance to the judge's judgment, etc. \\
with a score of 0-5 points; \\
s2:  Does it include the explanation of our viewpoint and the rebuttal of the other party's viewpoint. \\
Is our viewpoint reasonable? Is the rebuttal reasonable and well founded? Is there a correct \\
judgment on the current situation of the case? Whether it is conducive to providing judges with a \\
comprehensive and effective perspective, etc., score 0-5 points; \\
s3:  After a round of debate, can we give the outside world a comprehensive understanding of the case? \\
Can summarize the plot of the case, give the judge room to make judgments, etc., score 0-5 points \\
\textbf{Evaluation criteria}: \\
s1:  Did the content of the debate cite legal articles and cases, and is the relevance of legal articles \\
and cases to this case significant? Can you provide effective assistance to the judge's judgment.\\
0.5-1 points: Refers to some relevant laws and cases, which can provide reference for the judge's \\
judgment to a certain extent, but not comprehensive enough. \\
1-1.5 points: Some relevant laws and cases were cited, but it was not comprehensive or in-depth enough. \\
1.5-2.5 points: In addition to citing some legal articles and cases, the interpretation of applicability \\ 
is also relatively in-depth, which is more helpful for judges to make judgments. \\
2.5-4 points: Many relevant laws and cases have been cited, which has a high relevance to this case \\  
and a clear interpretation, providing effective
assistance for the judge's judgment. \\
4-5 points: Accurately cited a large number of highly relevant laws and cases, providing comprehensive, \\
in-depth, and highly valuable references for judges' judgments. \\
s2:  Does it include the explanation of our viewpoint and the rebuttal of the other party's viewpoint?\\
The viewpoint needs to be reasonable,  and the rebuttal needs to be well founded in order to provide the \\ 
judge with an effective perspective. \\
0-0.5 points: Explained our viewpoint and provided some rebuttal to the other party's viewpoint, but \\ 
there are deficiencies in the strength, rationality, and basis of the viewpoint and rebuttal. \\
0.5-1.5: We have clearly stated our viewpoint and have a certain degree of counterattack against \\
the other party, but the strength of the judge's decisive judgment is not strong enough. \\
1.5-3 points: Clearly and reasonably presented our viewpoint, and strongly refuted the other party's \\
viewpoint with sufficient evidence. \\
3-4.5 points: The explanation of our viewpoint has basically gained accurate and strong persuasiveness, \\ 
comprehensively elaborated our viewpoint, and provided most complete and reliable rebuttals to the \\
other party's viewpoint. \\
4.5-5 points: Accurately, deeply, and persuasively presented our viewpoint, and provided a comprehensive, \\ 
powerful, and well founded rebuttal to the other party's viewpoint. \\
s3:  After a round of debate, can we give the outside world a comprehensive understanding of the case? \\ 
Being able to summarize the plot of the case and provide the judge with room for judgment. \\
0-0.5 points: Described some of the case plot, but there are some key information omissions that \\ 
have certain limitations on the judge's judgment. \\
0.5-1.5 points: A comprehensive description of the case's plot can provide the outside world with a clear \\ 
understanding of the case and provide sufficient judgment space for the judge. \\
1.5-2.5 points: It basically covers the details of the case, and the outside world has a basic understanding \\ 
of the case. The judge can basically judge the outcome of the case. \\
2.5-3.5: Relatively complete, accurate, and comprehensive description of the case plot, providing \\ 
the outside world with a thorough understanding of the case and offering direction for the judge's \\
judgment. \\
3.5-5 points: A complete, accurate, and comprehensive description of the case plot provides the outside \\ 
world with a very thorough understanding of the case, providing judges with broad and sufficient \\
judgment space. \\
\bottomrule
\end{tabular}
\end{normalsize} 
\caption{The prompt for lawyer agent to evaluate its argument(translated from Chinese)}
\label{table10}
\end{table*}

\section{Generated Case}
\label{sec:appendix G}

A generated case is presented in Table \ref{table11}, which consists of case fact, plaintiff's indictment, and defendant's plead.

\begin{table*}[ht]
\centering
\begin{tabular}{ccp{0.6\textwidth}}
\hline
\textbf{Law Articles} & \textbf{Item} & \textbf{Content} \\
\hline
\multirow{3}{*}{Article 465} & Case fact & On August 15, 2023, the plaintiff Li signed a housing lease contract with the defendant Zhang, agreeing that Li would lease a property located in Haidian District, Beijing to Zhang for a period of one year, with a monthly rent of 5000 yuan. According to the contract, Zhang was required to pay the monthly rent before the 5th of each month. However, since October 2023, Zhang failed to pay the rent on time and Li repeatedly urged him to do so without success. Therefore, Li decided to file a lawsuit with the court, demanding the termination of the lease contract and the recovery of the overdue rent. \\
\cline{2-3}
 & plantiff's indictment & The plaintiff Li requests the court to order the termination of the lease contract between him and the defendant Zhang, and demands that Zhang pay a total of 15000 yuan in rent arrears (from October to December 2023), as well as bear the litigation costs of this case. The facts and reasons are: According to Article 509 of the Civil Code of the People's Republic of China, the parties to the contract should fully fulfill their obligations as agreed. The defendant Zhang failed to pay rent on time as agreed in the contract, which constitutes a breach of contract. The plaintiff has the right to terminate the contract and demand compensation for losses \\
\cline{2-3}
 & defendant's plead & The defendant Zhang argued that due to poor management of the company, the defendant is currently facing financial difficulties and is unable to pay rent temporarily. The defendant has communicated with the plaintiff and hopes to delay payment. They are also willing to make up for the overdue rent in one go after the Spring Festival. The defendant requests the court to consider the defendant's actual difficulties, and to give the defendant a lenient treatment and a certain grace period. \\
\hline
\end{tabular}
\caption{An example of a generated case(translated from Chinese).}
\label{table11}
\end{table*}

\end{document}